\documentclass[10pt,twocolumn,letterpaper]{article}

\usepackage{cvpr}              

\usepackage{caption} 

\usepackage{booktabs}
\usepackage{array}
\usepackage{multirow}
\usepackage{amsmath,amssymb}
\usepackage{pgfplots}
\pgfplotsset{compat=1.18}
\usepackage{tikz}
\usetikzlibrary{arrows.meta,positioning,calc}
\usepackage{xspace}
\usepackage{enumitem}

\newif\ifshowscaffold
\showscaffoldfalse
\definecolor{scaffold}{RGB}{200,0,0}
\definecolor{placeholderc}{RGB}{235,110,0}
\NewDocumentCommand{\goal}{+m}{\ifshowscaffold\par\smallskip\noindent
  \fcolorbox{scaffold}{scaffold!6}{\parbox{\dimexpr\linewidth-2\fboxsep-2\fboxrule\relax}{%
  \raggedright\footnotesize\color{scaffold}\textbf{Goal:} #1}}\par\smallskip\fi}
\NewDocumentCommand{\scaffnote}{+m}{\ifshowscaffold{\color{scaffold}\footnotesize\textbf{[Note:} #1\textbf{]}}\fi}
\NewDocumentCommand{\result}{+m}{#1}
\newcommand{\tbd}[1]{{\color{placeholderc}#1}}

\newcommand{\contentlegend}{\ifshowscaffold\par\noindent\fcolorbox{scaffold}{scaffold!6}{\parbox{\dimexpr\linewidth-2\fboxsep-2\fboxrule\relax}{%
  \raggedright\footnotesize\color{scaffold}\textbf{Legend.} Red: scaffolding (off: \texttt{\textbackslash showscaffoldfalse}).
  \tbd{Orange: not yet measured, or a choice to be confirmed} (expected demo values).
  {\color{black}Black: measured, derived, or method.}}}\par\medskip\fi}

\newcommand{\method}{CoDe\xspace}

\definecolor{cvprblue}{rgb}{0.21,0.49,0.74}
\usepackage[pagebackref,breaklinks,colorlinks,allcolors=cvprblue]{hyperref}

\def\confName{CVPR}
\def\confYear{2027}

\title{The Rhythm of Generated Video: Latent Chunking Leaves Periodic Traces in Compressed Bitstreams}

\author{
Chen Ye Gan$^{1}$ \qquad
Bryan Zhen Zhang$^{2}$ \qquad
Gene Wen$^{3}$ \qquad
Yuxing Han$^{1}$ \\
$^{1}$Tsinghua University \qquad
$^{2}$University of California, San Diego \\
$^{3}$New York University Shanghai
}

\begin{document}
\twocolumn[{%
\renewcommand\twocolumn[1][]{#1}%
\maketitle

\vspace{-3em}

\begin{center}
\begin{minipage}[b]{0.635\textwidth}
\centering
\includegraphics[width=\linewidth]{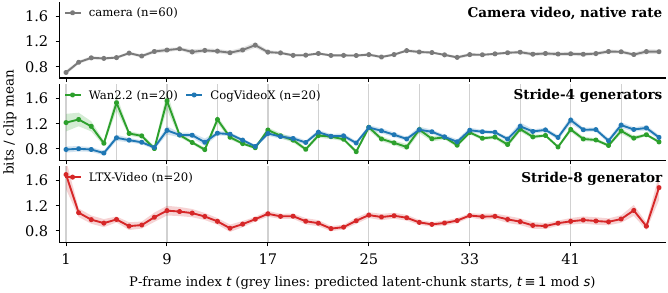}\\[-1pt]
{\scriptsize\textbf{(a)} Per-frame bits, x265 at constant QP 27, P-frames only}
\end{minipage}\hfill
\begin{minipage}[b]{0.335\textwidth}
\centering
\includegraphics[width=\linewidth]{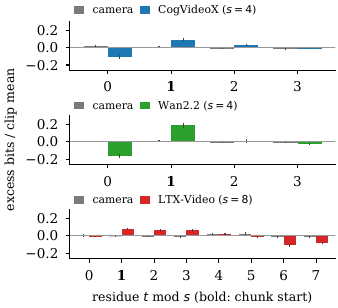}\\[-1pt]
{\scriptsize\textbf{(b)} Where in the chunk the extra bits fall}
\end{minipage}
\vspace{-0.5em}
\captionsetup{hypcap=false}
\captionof{figure}{\textbf{A standard rate-distortion optimizing encoder spends its bits on AI generated video in pulses.} 
(a) Bits per frame, divided by each clip's mean, for camera clips and for image-to-video clips generated from their first frames. The pulse comes every fourth frame for CogVideoX and Wan2.2 and every eighth for LTX-Video, the temporal strides $s$ of their autoencoders; grey lines mark the predicted latent-chunk starts $t\equiv1 \pmod s$. Camera footage at its native frame rate has no pulse.
(b) The same bits, detrended and averaged by frame residue $t \bmod s$ over P-frames 6--45: the stride-4 models put their excess on the chunk-start residue.}
\label{fig:teaser}
\end{center}
\vspace{0.4em}
}]
\begin{abstract}
Most AI-generated video detectors analyze decoded pixels and overlook how videos are produced. We show that causal video autoencoders leave an architecture-specific rhythm in compressed bitstreams: their temporal stride concentrates "surprise" at latent-chunk boundaries, so rate--distortion optimized encoders allocate bits periodically. We observe four-frame pulses in CogVideoX and Wan2.2 and eight-frame pulses in LTX-Video, which are absent from similar camera footage. Passing camera footage through a generator’s autoencoder alone induces the pulse. A training-free peak-ratio statistic separates generated from camera video (AUC 0.84--0.94) and persists across tested quantization settings and H.264/H.265 encoding. Frame-size traces also identify four commercial generators, with 89.2\%, 84.1\%, and 83.0\% accuracy under H.264, H.265, and AV1, beating the pixel-based UnivFD baseline (72.7--73.6\%). Our results directly link generative architecture to measurable coding demand.
\end{abstract}

\contentlegend
    
\section{Introduction}
\label{sec:intro}

\goal{Phenomenon first (\cref{fig:teaser}), then the proposed mechanism (surprise enters once per latent chunk), then the story: the pattern in the latent representation, its survival through video encoding, its robustness under the tested settings, and small classifiers that read it from the bitstream.}

Text-to-video and image-to-video generators now produce clips that are hard to tell from camera footage by eye~\cite{ho2022video,blattmann2023stable,sora2024,wan2025}.
Much work detects AI-generated images, and increasingly video, from decoded pixels~\cite{wang2020cnn,ojha2023towards,vahdati2024beyond}.
Far less links detection to the generative process itself, perhaps because a delivered video is shaped not only by the generator but also by a video encoder, a system many vision researchers rarely look inside.
This paper follows a generated video through both.

Nearly all current generators are \emph{latent} generators: a diffusion or flow model samples a compact latent video, and a video autoencoder decodes it to pixels~\cite{yang2025cogvideox,wan2025,hacohen2024ltx}.
To compress time, these autoencoders are causal 3-D networks with a fixed temporal stride $s$: the first frame is decoded alone, and every later latent frame is decoded into $s$ pixel frames.
Call the part of a frame that cannot be predicted from its past its \emph{surprise}.
In camera footage, surprise arrives continuously, at roughly the same rate in every frame.
We propose that in a latent video it can enter only where a new latent frame enters, once every $s$ frames, since the frames in between are decoded from the same latent.
Generated video should then carry a surprise rhythm with period $s$, locked in phase to the latent-chunk grid (\cref{fig:rhythm}).
The stride is an established architectural fact; the rhythm is the mechanism we propose and test.

The second half of the pipeline is a conventional video encoder.
Codecs such as H.264/AVC~\cite{wiegand2003overview}, HEVC~\cite{sullivan2012overview} and AV1 predict each frame from frames already coded and are rate--distortion (RD) optimized~\cite{sullivan1998rate}: they spend bits only on what their temporal prediction cannot anticipate, that is, on surprise (\cref{fig:encode}).
The bits an encoder spends on each frame, readable from the bitstream without decoding, should therefore trace the generator's rhythm.
\Cref{fig:teaser} shows that they do.
\result{Re-encoded under identical settings, clips from CogVideoX~\cite{yang2025cogvideox} and Wan2.2~\cite{wan2025} (stride 4) pulse every fourth frame and clips from LTX-Video~\cite{hacohen2024ltx} (stride 8) every eighth, while content-matched camera clips do not pulse; a training-free peak ratio separates them at an AUC of 0.84--0.94.}

We tell the story in three steps.
First, the pattern originates in the latent representation: \result{the pulse frequency moves with the stride, its phase follows the chunk grid for the stride-4 models, and passing camera clips through a generator's autoencoder alone creates it} (\cref{sec:view,sec:causal}).
Second, it survives video encoding: \result{in both H.264 and H.265, and from quantization parameter (QP) 22 to 37 (AUC 0.92 to 0.80)} (\cref{sec:encoding,sec:survival}).
Third, its persistence has limits we state plainly: \result{delivery-style encodes with rate control and B-frames erase the per-frame pulse in 14 of 15 tested cases, although x265 still places its reference frames on the stride-4 grid}.

The traces are simple to read (\cref{sec:detectors}).
We divide each frame's size by the clip's mean, giving frame-size ratios, and feed them to a logistic regression or to a small transformer with 26K parameters.
We evaluate on DeepAction~\cite{bohacek2024deepaction}, 2{,}605 clips of human actions from camera footage and six generators, and on a self-collected \emph{Frontier} dataset of four commercial generators (Grok, HeyGen, MiniMax, Wan) on 40 shared prompts, with 160, 159 and 155 clips re-encoded in H.264, H.265 and AV1.
\result{On Frontier, where all generators share frame size and length, the logistic regression identifies the generator at 89.2\%, 84.1\% and 83.0\% in H.264, H.265 and AV1, and the transformer at 80.6\% and 78.4\% in H.264 and H.265 (chance 25\%); on the same clips and task, the pixel-based UnivFD detector~\cite{ojha2023towards} reaches 72.7\%, 73.6\% and 73.4\%, so bitstream traces outperform it by 10--17 points.}
\result{The advantage is specific to this generator-identification task: for camera versus generated video on DeepAction with size and length controlled, UnivFD remains stronger (90.5\% vs.\ 69.9--74.1\% balanced accuracy on 16-frame crops, known generators), while gradient-boosted trees on bit statistics reach 88.8\% on 192-frame clips.}
On DeepAction we equalize frame size and clip length throughout, since its sources differ in file format, and a classifier that works need not use the rhythm, which we test with ablations and attributions (\cref{sec:main,sec:posthoc}).

Our contributions are two:
\begin{itemize}[leftmargin=*,itemsep=1pt,topsep=2pt]
\item \textbf{Linking temporal compression in video generation to a detectable trace.}
To our knowledge, we are the first to tie a video generator's temporal stride to a measurable trace, as prior work did for spatial up-sampling and autoencoders in images~\cite{frank2020leveraging,durall2020watch,ricker2024aeroblade}.
\item \textbf{A decoding-free method that reads the rhythm directly from the encoded bitstream.}
\method takes frame and row sizes from the bitstream without reconstructing pixels and \result{identifies four frontier commercial generators at 83.0--89.2\%, against 72.7--73.6\% for the pixel-based UnivFD}
\end{itemize}

\section{Background and Related Work}
\label{sec:related}
\goal{Give only the background the unifying view needs: pixel-domain detectors (what we are not), latent generators and their temporal stride (the cause), RD coding (the instrument), and compressed-domain analysis (closest prior work; the authors' journal paper cited in the third person).}

\paragraph{AI-generated image and video detection.}
CNN-generated images carry traces that a classifier trained on one generator can detect in others~\cite{wang2020cnn}, and frequency analyses attribute many of them to up-sampling layers~\cite{frank2020leveraging,durall2020watch}.
Large pretrained features generalize better to unseen generators~\cite{ojha2023towards}, diffusion images leave weaker traces~\cite{corvi2023detection}, and the reconstruction error of a latent model's autoencoder exposes images it generated~\cite{ricker2024aeroblade}.
B-Free~\cite{guillaro2025bias} removes content, format and resize biases one at a time; we adopt the same discipline for video (\cref{sec:main}).
For video, synthetic-image detectors fail on synthetic video, compression is not the main reason, and video generators leave traces a learned detector finds even after H.264 re-compression~\cite{vahdati2024beyond}.
Spatio-temporal detectors trained on GenVideo~\cite{chen2024demamba} and training-free detectors that model the likelihood of real video~\cite{benhayun2026stall} extend this line.
All of these operate on decoded pixels; we instead analyze how a video encoder allocates its bits, read from the bitstream without reconstructing pixels.

\paragraph{Latent video generators and temporal stride.}
CogVideoX~\cite{yang2025cogvideox} and Wan~\cite{wan2025} use causal 3-D autoencoders with temporal stride $s{=}4$, in which the first frame is encoded alone and every later latent frame summarizes four pixel frames; LTX-Video~\cite{hacohen2024ltx} compresses time by $s{=}8$.
The stride is an architectural constant, shared by every clip a model produces, whatever its content or sampler.
Reconstruction by a latent model's autoencoder already exposes generated \emph{images}~\cite{ricker2024aeroblade}; we study the effect of the autoencoder's temporal structure on the generated video, and use the resulting rhythm for classification.

\paragraph{Video coding and rate--distortion optimization.}
Hybrid codecs such as H.264/AVC~\cite{wiegand2003overview} and HEVC~\cite{sullivan2012overview} predict each block from the same frame (intra) or from reference frames (inter) and code the residual.
Frames form groups of pictures (GOPs) that begin with an intra-coded I-frame, followed by predicted P- and B-frames.
For each block the encoder minimizes $J = D + \lambda R$~\cite{sullivan1998rate}, with $\lambda$ set by the quantization parameter (QP).
The resulting allocation is an operational estimate of the content's rate--distortion function~\cite{shannon1959coding} as seen through the encoder's predictors, and when frames depend on each other the optimal allocation favors heavily referenced frames~\cite{ramchandran1994bit}.
Rate control (constant QP, constant rate factor (CRF), target bitrate), adaptive quantization and MB-tree then move bits in space and time, which is why the encoder and its settings are part of our measurement.
Holding these settings fixed (constant QP, one I-frame followed by P-frames, adaptive tools off) removes the encoder's own patterns, so the bit allocation that remains can be studied as a property of the content (\cref{sec:experiments}).

\paragraph{Compressed-domain analysis.}
Motion vectors and residuals support action recognition without full decoding~\cite{wu2018compressed}, and entropy-based compression statistics were recently shown to form a rich and highly efficient spatial--temporal representation for video understanding~\cite{anon2026entropy}.
\scaffnote{Double-blind: \cite{anon2026entropy} is the authors' own journal paper; cite in the third person only and restore the full reference at camera-ready.}
These works recognize \emph{what} a video shows.
Closest to us, codec motion-vector statistics have been used to rate the temporal realism of generated video~\cite{cakiroglu2025temporal}; that study reports oscillations in per-frame averages but attributes them to interpolation and resamples clips to a common length, which would blur a native stride rhythm. 
The same group showed that frame-type counts identify which corpus a clip came from~\cite{cakiroglu2026corpus}, the confound our protocol is built to remove.
We ask what the encoder's bit allocation reveals about \emph{how} a video was made; to our knowledge, a stride-locked rhythm in encoder bits has not been reported before.

\section{Generator-Induced Temporal Rhythms}
\label{sec:view}
\goal{The generator's side, in ML terms: a causal video autoencoder of stride $s$ lets new information in once per latent chunk, so generated video has a periodic surprise rhythm with two testable signatures (period $=s$, phase $=$ chunk grid); camera video is the member of the family with no chunking.}

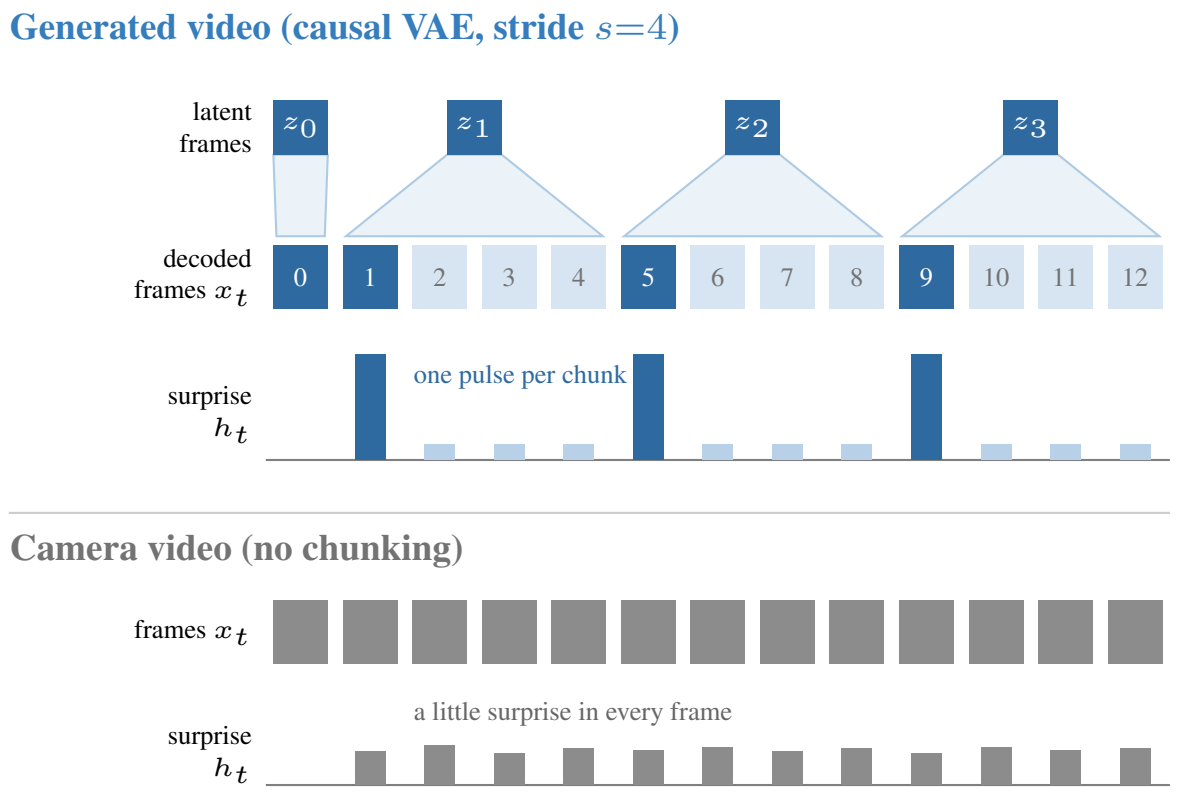
\begin{figure}[t]
\centering
\begin{tikzpicture}[x=1cm,y=1cm,font=\scriptsize,>={Stealth[length=3.5pt]},
  lab/.style={anchor=east,font=\tiny,align=right,inner sep=1pt},
  hd/.style={anchor=west,font=\scriptsize\bfseries,inner sep=0pt}]
\def\px{0.46}\def\fw{0.36}\def\xo{1.75}
\node[hd,text=cvprblue] at (0,4.15) {Generated video (causal VAE, stride $s{=}4$)};
\foreach \k/\a/\b in {0/0/0,1/1/4,2/5/8,3/9/12}{
  \pgfmathsetmacro\xa{\xo+\px*\a}\pgfmathsetmacro\xb{\xo+\px*\b+\fw}
  \pgfmathsetmacro\xc{(\xa+\xb)/2}
  \fill[cvprblue!10,draw=cvprblue!40] (\xa+0.02,2.78) -- (\xb-0.02,2.78) -- (\xc+0.18,3.32) -- (\xc-0.18,3.32) -- cycle;
  \fill[cvprblue!85!black] (\xc-0.18,3.32) rectangle (\xc+0.18,3.68);
  \node[text=white,font=\tiny] at (\xc,3.5) {$z_{\k}$};}
\node[lab] at (\xo-0.1,3.5) {latent\\frames};
\foreach \t in {0,...,12}{\fill[cvprblue!20] ({\xo+\px*\t},2.3) rectangle ++(\fw,0.42);}
\foreach \t in {0,1,5,9}{\fill[cvprblue!85!black] ({\xo+\px*\t},2.3) rectangle ++(\fw,0.42);}
\foreach \t in {2,3,4,6,7,8,10,11,12}{\node[font=\tiny,text=black!55] at ({\xo+\px*\t+\fw/2},2.51) {\t};}
\foreach \t in {0,1,5,9}{\node[font=\tiny,text=white] at ({\xo+\px*\t+\fw/2},2.51) {\t};}
\node[lab] at (\xo-0.1,2.51) {decoded\\frames $x_t$};
\draw[black!50] (\xo-0.05,1.3) -- ({\xo+\px*12+\fw+0.05},1.3);
\foreach \t in {1,...,12}{\fill[cvprblue!35] ({\xo+\px*\t+0.08},1.3) rectangle ++(0.2,0.1);}
\foreach \t in {1,5,9}{\fill[cvprblue!85!black] ({\xo+\px*\t+0.08},1.3) rectangle ++(0.2,0.7);}
\node[lab] at (\xo-0.1,1.6) {surprise\\$h_t$};
\node[font=\tiny,text=cvprblue!85!black,anchor=west] at ({\xo+\px*1.75},1.85) {one pulse per chunk};
\draw[black!20] (0,0.95) -- ({\xo+\px*12+\fw+0.05},0.95);
\node[hd,text=black!55] at (0,0.7) {Camera video (no chunking)};
\foreach \t in {0,...,12}{\fill[black!45] ({\xo+\px*\t},-0.05) rectangle ++(\fw,0.42);}
\node[lab] at (\xo-0.1,0.16) {frames $x_t$};
\draw[black!50] (\xo-0.05,-0.85) -- ({\xo+\px*12+\fw+0.05},-0.85);
\foreach \t/\h in {1/0.22,2/0.26,3/0.21,4/0.24,5/0.23,6/0.25,7/0.22,8/0.24,9/0.21,10/0.25,11/0.23,12/0.24}{
  \fill[black!45] ({\xo+\px*\t+0.08},-0.85) rectangle ++(0.2,\h);}
\node[lab] at (\xo-0.1,-0.65) {surprise\\$h_t$};
\node[font=\tiny,text=black!60,anchor=south west] at ({\xo+\px*1.75},-0.58) {a little surprise in every frame};
\end{tikzpicture}
\vspace{-0.5em}
\caption{\textbf{A causal video autoencoder lets new information in once per chunk.}
The decoder expands $z_0$ into frame 0 and each later $z_k$ into $s$ frames; only frame 0 and the first frame of each chunk (dark; $t\equiv1 \pmod s$) carry much that the past cannot predict.
Camera footage carries surprise at a roughly constant rate.}
\label{fig:rhythm}
\vspace{-0.6em}
\end{figure}

\paragraph{Latent video generation.}
Nearly all current video generators work in a latent space.
A sampler, a diffusion or flow-matching transformer, turns noise into a short sequence of latent frames $z_0, z_1, \dots$.
A causal 3-D variational autoencoder (VAE) then decodes them to pixels.
Its decoder is \emph{causal} (it uses only $z_{\le k}$ to produce chunk $k$) and has a temporal \emph{stride} $s$: $z_0$ decodes to frame $x_0$ alone, and each later $z_k$ decodes to a chunk of $s$ frames, $x_{(k-1)s+1},\dots,x_{ks}$ (\cref{fig:rhythm}).
The stride is $s{=}4$ for CogVideoX~\cite{yang2025cogvideox} and Wan2.2~\cite{wan2025} and $s{=}8$ for LTX-Video~\cite{hacohen2024ltx}.

\paragraph{In an autoencoder, surprise enters once per chunk.}
\label{sec:family}
The \emph{surprise} of a frame is the part of it that cannot be predicted from its past.
Formally, for a video $x_0,\dots,x_{T-1}$ it is the conditional entropy $h_t = H(x_t \mid x_{<t})$.
In camera footage the scene and the sensor change a little in every frame, so $h_t$ is roughly constant in $t$.

We propose that in a latent video, new information can enter only with a new latent frame.
The later frames of a chunk are decoded from the same $z_k$ as its first, so most of the chunk's surprise should sit on its first frame:
\begin{equation}
\mathbb{E}[h_t] \approx \bar h + a\,\mathbf{1}[\,t \bmod s = 1\,].
\label{eq:periodic}
\end{equation}
Under this model the surprise is periodic in $s$, and its phase is fixed by the chunk grid, not by the content; \cref{sec:experiments} tests both predictions.

\Cref{eq:periodic} makes two testable predictions that can be checked separately.
\emph{Stride}: the period moves with $s$, a dose-response test between stride-4 and stride-8 generators.
\emph{Phase}: the pulse is locked to the chunk grid, so removing the first $k$ frames of a clip rotates it by $k$.
The model is idealized: temporal up-sampling that mixes neighbouring latents can blur the phase, and \result{LTX-Video's phase is only partly locked} (\cref{sec:exist}).

The sources then form a family: camera video is the member with no chunking, $a{=}0$.
Generators with stride 4 and stride 8 are two further members.
Stride and phase index the family, much as the pooling axis indexes normalization layers~\cite{wu2018group}.

\section{Encoder-Induced Bit Allocation Rhythms}
\label{sec:encoding}
\goal{The encoder's side, in systems terms: a rate--distortion optimized encoder at fixed QP pays P-frame bits for surprise, so its per-frame bits (coding demand, read from the bitstream without decoding) trace the rhythm of \cref{sec:view}; a clean protocol is needed because the encoder's own structure (B-frames, keyframes, rate control, frame dropping) can mask or mimic it.}

\begin{figure}[t]
\centering
\begin{tikzpicture}[x=1cm,y=1cm,font=\scriptsize,>={Stealth[length=3.5pt]},
  blk/.style={draw=black!70,fill=white,rounded corners=2pt,align=center,inner sep=2pt,font=\tiny},
  note/.style={font=\tiny,align=center,inner sep=1pt}]
\node[blk,minimum height=0.6cm] (x) at (0.35,2.2) {frame\\$x_t$};
\node[draw,circle,inner sep=0.6pt,fill=white] (sum) at (1.55,2.2) {\tiny$-$};
\node[blk] (mc) at (1.55,3.15) {predict from\\previous frame};
\node[blk,minimum height=0.6cm] (q) at (4.2,2.2) {quantize at\\fixed QP, code};
\node[blk,minimum height=0.6cm] (b) at (6.95,2.2) {bits for\\frame $t$: $b_t$};
\draw[->] (x) -- (sum);
\draw[->] (mc) -- (sum);
\draw[->] (sum) -- node[below,note,yshift=-1pt]{residual\\$\approx$ surprise} (q);
\draw[->] (q) -- (b);
\def\bx{1.0}\def\bp{0.19}
\node[note,anchor=west,font=\tiny\bfseries] at (0,1.35) {bits per frame $b_t$};
\draw[black!50] (\bx-0.05,0.3) -- (\bx+3.05,0.3);
\foreach \t/\h in {1/0.75,2/0.22,3/0.19,4/0.21,5/0.73,6/0.23,7/0.18,8/0.2,9/0.74,10/0.21,11/0.19,12/0.22,13/0.72,14/0.2,15/0.21,16/0.19}{
  \fill[cvprblue!75] ({\bx+\bp*(\t-1)},0.3) rectangle ++(0.14,\h);}
\draw[black!50] (\bx-0.05,-0.55) -- (\bx+3.05,-0.55);
\foreach \t/\h in {1/0.3,2/0.28,3/0.32,4/0.29,5/0.27,6/0.31,7/0.3,8/0.28,9/0.31,10/0.29,11/0.3,12/0.27,13/0.31,14/0.29,15/0.3,16/0.28}{
  \fill[black!45] ({\bx+\bp*(\t-1)},-0.55) rectangle ++(0.14,\h);}
\node[note,text=cvprblue,anchor=east] at (\bx-0.1,0.5) {generated};
\node[note,text=black!55,anchor=east] at (\bx-0.1,-0.4) {camera};
\draw[->,black!70] (4.6,-0.55) -- (7.95,-0.55) node[above left,note]{$f$};
\draw[->,black!70] (4.65,-0.6) -- (4.65,1.15);
\node[note,anchor=west,font=\tiny\bfseries] at (4.55,1.35) {spectrum of $b_t$};
\draw[cvprblue!85!black,thick,line join=round] plot coordinates {(4.65,-0.35)(4.9,-0.4)(5.15,-0.34)(5.4,-0.39)(5.65,-0.35)(5.9,-0.38)(6.05,-0.36)(6.15,0.85)(6.25,-0.36)(6.4,-0.39)(6.65,-0.35)(6.9,-0.39)(7.15,-0.35)(7.4,-0.38)(7.55,-0.36)(7.65,0.35)};
\draw[black!50,thick,line join=round] plot coordinates {(4.65,-0.44)(4.9,-0.48)(5.15,-0.43)(5.4,-0.47)(5.65,-0.44)(5.9,-0.48)(6.15,-0.44)(6.4,-0.47)(6.65,-0.43)(6.9,-0.47)(7.15,-0.44)(7.4,-0.48)(7.65,-0.45)};
\draw[black!40,dotted] (6.15,-0.55) -- (6.15,0.85);
\node[note,anchor=north] at (6.15,-0.58) {$1/s$};
\node[note,anchor=north] at (7.65,-0.58) {$2/s$};
\node[note,text=cvprblue!85!black,anchor=west] at (6.25,0.75) {peak (generated)};
\node[note,text=black!55,anchor=west] at (6.7,-0.2) {flat (camera)};
\draw[black!20] (0,1.62) -- (7.95,1.62);
\end{tikzpicture}
\vspace{-0.5em}
\caption{\textbf{At fixed QP, a P-frame's bits measure its surprise.}
The encoder predicts each frame from the previous one and pays only for the residual.
On generated video the bits per frame $b_t$ pulse once per chunk and their spectrum peaks at $1/s$, with harmonics; on camera video they are flat.
\result{Measured: median peak ratio 27--29 for the stride-4 generators, 1.1--1.4 for camera clips} (\cref{tab:stride}).}
\label{fig:encode}
\vspace{-0.6em}
\end{figure}
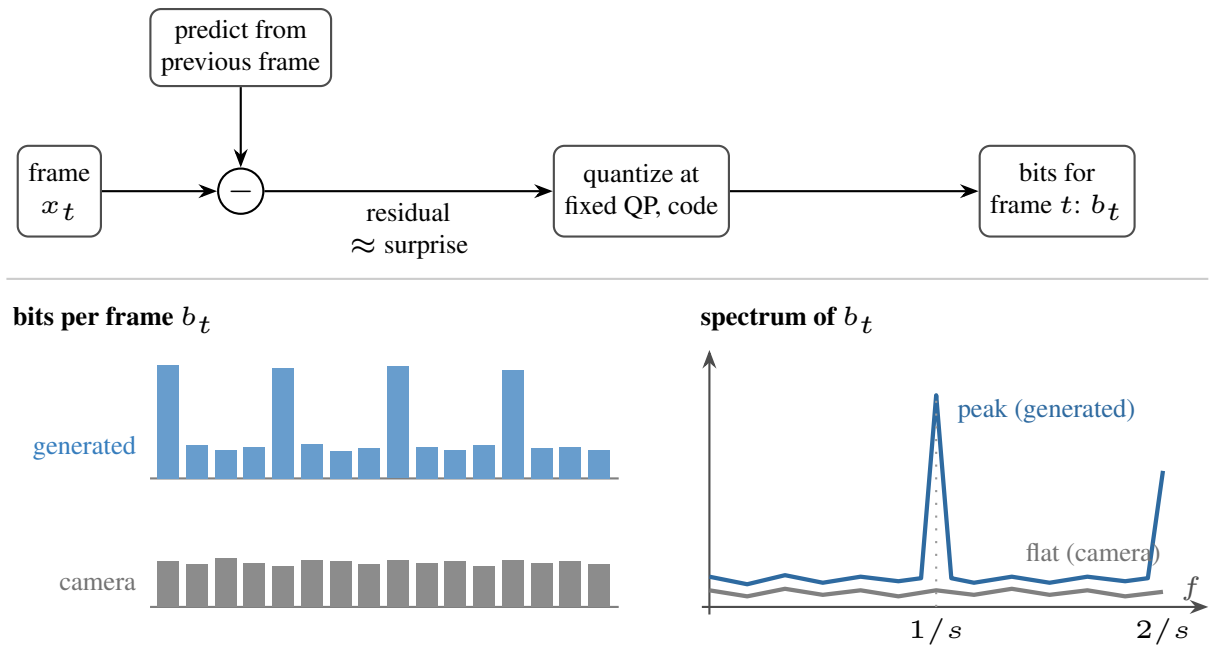

\paragraph{How an encoder spends bits.}
A modern encoder such as HEVC~\cite{sullivan2012overview} codes a video as a group of pictures (GOP).
An intra (I) frame is coded on its own, like a still image.
A predicted (P) frame is coded as a prediction from an earlier, already decoded frame (motion-compensated copying of blocks) plus a \emph{residual}, the part the prediction got wrong.
A bidirectional (B) frame may also predict from a later frame.
The residual is quantized with a step size set by the quantization parameter (QP; larger means coarser) and entropy-coded.
The encoder is rate--distortion (RD) optimized: for each block it chooses the mode, motion and quantized residual that minimize $J = D + \lambda(\mathrm{QP})\,R$, distortion plus $\lambda$ times rate~\cite{sullivan1998rate}.
At a fixed QP, what a P-frame costs is therefore what its predictor could not anticipate, \ie its surprise (\cref{fig:encode}).

\paragraph{Coding demand.}
\label{sec:def}
HEVC splits each frame into coding tree units (CTUs), square blocks of up to $64{\times}64$ pixels.
We record the bits the encoder $\mathcal{E}$ spends on each block $(i,j)$ of frame $t$, $B_t(i,j)$, counting mode, motion and residual syntax, and call this allocation the video's \emph{coding demand}.
We use it at three resolutions, $b_t$, $r_t(i)$ and $B_t(i,j)$,
per frame, per row of blocks, and per block:
\begin{equation}
b_t = \textstyle\sum_{i,j} B_t(i,j), \qquad r_t(i) = \sum_j B_t(i,j).
\label{eq:demand}
\end{equation}
In a P-frame at fixed QP, $b_t$ is an operational estimate of $h_t$ as seen through the encoder's predictors, so coding demand measures surprise.
We divide by the clip mean, $\tilde b_t = b_t/\bar b$, to keep only the shape of the allocation.

\paragraph{Read without decoding.}
None of these quantities requires reconstructing a pixel.
The frame sizes $b_t$ are the packet sizes in the container.
With HEVC wavefront parallel processing (WPP), each row of CTUs is coded as its own substream so that rows can be decoded in parallel, and the slice header lists where each substream starts; the differences between these entry points give $r_t(i)$.

\section{A Decoding-Free Approach to Reading the Rhythm Directly from the Bitstream}
\label{sec:detectors}
\goal{The method: parse frame and row sizes from the bitstream (no decoding), normalize per clip, and read them with three detectors of increasing capacity: the training-free peak ratio and phase, a 26k-parameter transformer on row bytes, and gradient-boosted trees on about 50 bit statistics.}

\begin{figure}[t]
\centering
\begin{tikzpicture}[x=1cm,y=1cm,font=\tiny,>={Stealth[length=3pt]},
  st/.style={draw=black!70,fill=white,rounded corners=2pt,align=center,inner sep=2pt,minimum height=0.62cm,font=\tiny},
  det/.style={draw=cvprblue!80,fill=cvprblue!8,rounded corners=2pt,align=center,inner sep=1.5pt,minimum width=1.5cm,minimum height=0.36cm,font=\tiny}]
\node[st] (bs) at (0.55,0) {bitstream\\(HEVC)};
\node[st] (pa) at (2.0,0) {parse packet\\\& row sizes\\(no decoding)};
\node[st] (no) at (3.5,0) {normalize\\per clip\\$\tilde b_t$, $\tilde r_t(i)$};
\node[det] (d2) at (5.25,0) {(b) transformer};
\node[det] (d1) at (5.25,0.5) {(a) $\mathrm{PR}_s$ \& phase};
\node[det] (d3) at (5.25,-0.5) {(c) trees};
\node[st] (out) at (7.25,0) {real / generated\\or generator ID};
\draw[->] (bs) -- (pa);
\draw[->] (pa) -- (no);
\foreach \d in {d1,d2,d3}{\draw[->] (no.east) -- (\d.west);\draw[->] (\d.east) -- (out.west);}
\end{tikzpicture}
\vspace{-0.5em}
\caption{\textbf{\method reads frame and row sizes from the bitstream and never decodes pixels.}
The three detectors share the same normalized coding-demand signals: (a) needs no training, (b) and (c) are learned.}
\label{fig:pipeline}
\vspace{-0.6em}
\end{figure}
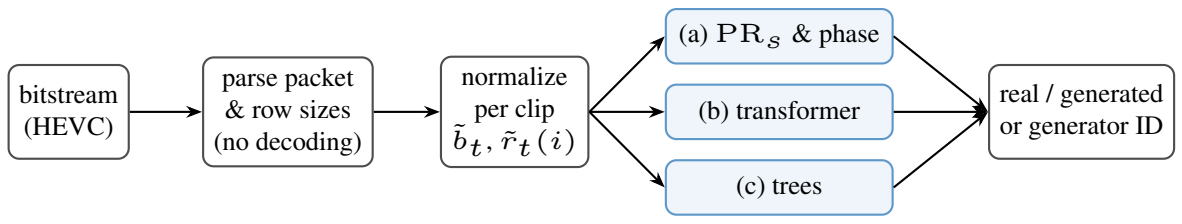

Our method, \method, reads the rhythm directly from the encoded bitstream and never reconstructs a pixel (\cref{fig:pipeline}).
A clip is coded at constant QP with one I-frame followed only by P-frames (\cref{sec:experiments}); a delivered file is first re-encoded.
A parser reads the packet sizes $b_t$ and the WPP entry points $r_t(i)$ without decoding a single pixel.
The signals are normalized per clip and passed to one of three detectors.

\paragraph{(a) Peak ratio and phase, training-free.}
In words, the peak ratio asks how far the spectrum of the bits stands out at the stride frequency compared with nearby frequencies.
We take P-frames 6--45 (skipping the transient after the I-frame), detrend $\tilde b_t$ linearly, and compute its power spectrum $P(f)$ over these $N$ frames.
For a candidate stride $s$,
\begin{equation}
\mathrm{PR}_s = \frac{P(1/s)}{\operatorname{median}_{0<|m|\le 4} P(1/s + m/N)},
\label{eq:pr}
\end{equation}
the power at $1/s$ over the median of its eight neighbouring bins, harmonics excluded.
A flat spectrum gives $\mathrm{PR}_s\approx1$.
The \emph{phase} is the residue $t \bmod s$ that receives the most bits; \cref{eq:periodic} predicts residue 1.
We score real versus generated by the area under the ROC curve (AUC), the probability that a generated clip outscores a camera clip.

\paragraph{(b) A lightweight transformer on row bytes.}
The row sequence $r_t(i)$ is cut into 16-frame patches per row.
A linear layer embeds each patch, and a sinusoidal row position is added.
Rows are pooled by mean and max, and a 2-layer transformer (width 32, 4 heads, feed-forward 64) with a learned temporal position runs over the patches, followed by mean pooling and a linear head.
It has 26{,}247 parameters for seven classes (26{,}148 for four), of which 6{,}016 are the temporal position table; values are z-scored over the training set.

\paragraph{(c) Gradient-boosted trees on bit statistics.}
From $\tilde b_t$ and $r_t(i)$, normalized per frame type, we compute about 50 statistics: quantiles and moments, band powers of the spectrum, $\mathrm{PR}_4$, $\mathrm{PR}_8$ and phase, temporal autocorrelation, and the spatial concentration of the rows.
They are fed to gradient-boosted decision trees (GBDT)~\cite{ke2017lightgbm}, an ensemble of small trees fitted one after another to the remaining error.

\section{Experiments}
\label{sec:experiments}
\goal{Test the view in the order it makes predictions: the rhythm exists and differs by family (\ref{sec:exist}); the autoencoder causes it (\ref{sec:causal}); it propagates into the encoded bitstream (\ref{sec:survival}); whether classifiers built on it work once format is controlled (\ref{sec:main}); what they actually use (\ref{sec:posthoc}); where in the frame it lives (\ref{sec:pending}).}

\paragraph{Data.}
Throughout, $\mathrm{PR}_s$ is the peak ratio of the per-frame bit spectrum at frequency $1/s$ for stride $s$ (\cref{sec:detectors}), and AUC the area under the ROC curve for camera vs.\ generated clips.
The \emph{stride corpus} isolates generation.
It has 20 camera clips of natural scenes from the Xiph derf 1080p collection~\cite{xiphderf} and, for each, an image-to-video clip generated from its first frame by CogVideoX-5B-I2V~\cite{yang2025cogvideox} ($720{\times}480$, 30 steps), Wan2.2-TI2V-5B~\cite{wan2025} ($1280{\times}704$, 50 steps) and LTX-Video~\cite{hacohen2024ltx} ($704{\times}480$, 50 steps), all 49 frames, seed 0.
Each generated clip has a content-matched camera reference: the first 49 frames of its source at the native frame rate, scaled and cropped to the generator's resolution, with no frame dropped.
Clips are encoded from lossless frames by x265 and x264 at a constant quantization parameter (QP), with one I-frame followed only by P-frames and with adaptive quantization, MB-tree and scene-cut detection off, so that the encoder adds no rhythm of its own.
The self-collected \emph{Frontier} set has clips of four commercial generators (Grok, HeyGen, MiniMax, Wan) for 40 shared prompts, re-encoded at $832{\times}480$ and 192 frames by x264 (constant QP 23), x265 (constant QP 28, wavefront parallel processing (WPP), 8 rows) and an AV1 encoder, all with keyframes every 16 frames; there is no camera class, and only Wan differs in format (30 instead of 24\,fps).
AV1 uses libaom's real-time mode, because its good-quality mode imposes a 4/8/16-frame cadence even on camera clips.
\emph{DeepAction}~\cite{bohacek2024deepaction} has 2{,}605 clips of human actions from Pexels (camera) and six generators, re-encoded by x265 with WPP at constant QP 28, keyframes every 16 frames and no B-frames.
With WPP, x265 also exposes the bytes of each row of coding blocks.
All splits are prompt-grouped: no prompt appears in both training and test data.

\paragraph{The rhythm exists and differs by family.}
\label{sec:exist}

\begin{table}[t]
\centering
\footnotesize
\setlength{\tabcolsep}{2.6pt}
\begin{tabular}{@{}lccccc@{}}
\toprule
& & Median & \multicolumn{2}{c}{Share of clips} & \\
\cmidrule(lr){4-5}
Generator & $s$ & $\mathrm{PR}_s$ & $\mathrm{PR}_s{>}3$ & $\mathrm{PR}_s{>}10$ & AUC \\
\midrule
CogVideoX & 4 & \textbf{27.3}\,/\,1.4 & 0.80\,/\,0.25 & 0.65\,/\,0.05 & 0.84 \\
Wan2.2    & 4 & \textbf{29.0}\,/\,1.1 & 0.85\,/\,0.15 & 0.65\,/\,0.00 & 0.94 \\
LTX-Video & 8 & \textbf{6.1}\,/\,1.1  & 0.75\,/\,0.10 & 0.40\,/\,0.00 & 0.89 \\
\bottomrule
\end{tabular}
\vspace{-0.4em}
\caption{\textbf{Generated clips pulse at their own stride; camera clips do not.}
x265, QP 27. Pairs are generated\,/\,camera: 20 generated clips per row vs.\ their 20 content- and resolution-matched camera references.
$\mathrm{PR}_s$: peak ratio of the bit spectrum at the generator's own stride.}
\label{tab:stride}
\vspace{-0.8em}
\end{table}

\begin{figure}[t]
\centering
\includegraphics[width=\linewidth]{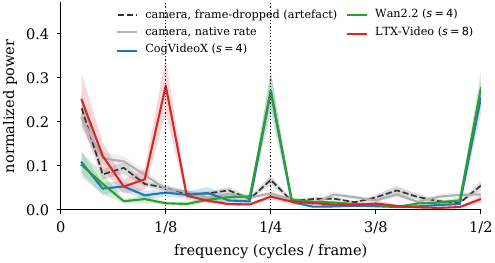}
\vspace{-1.6em}
\caption{\textbf{The peak follows the stride.} Mean ($\pm$ standard error, s.e.) power spectrum of normalized, detrended per-frame bits, x265 QP 27, P-frames 6--45. Dashed: camera clips frame-dropped to the generator's rate, which acquire a spurious peak at $1/4$; all camera references in this paper are therefore used at their native rate.}
\label{fig:spectrum}
\vspace{-0.8em}
\end{figure}

Does generated video carry a bit rhythm at the autoencoder's stride, and camera video not?
\result{Both stride-4 generators peak at exactly $f{=}1/4$, with a harmonic at $1/2$, and the stride-8 generator at exactly $1/8$; camera clips at their native rate show no peak} (\cref{fig:spectrum}).
\result{The median peak ratio at the generator's own stride is 27--29 for CogVideoX and Wan2.2 and 6.1 for LTX-Video, against 1.1--1.4 for their camera references (\cref{tab:stride}). The AUC is 0.84--0.94 with x265 and 0.85--0.93 with x264 (CogVideoX, Wan2.2, LTX-Video: 0.85, 0.90, 0.93).}
Generated video does not have a flatter bit spectrum than camera video; it concentrates the temporal variation of its bits at the stride frequency.
\result{Doubling the stride moves the peak from $1/4$ to $1/8$: LTX-Video has $\mathrm{PR}_8{=}6.1$ but $\mathrm{PR}_4{=}2.4$, whereas the stride-4 generators have $\mathrm{PR}_8$ of 0.6--1.4.}
The scenes are shared across generators, so only the stride explains the shift.
If the rhythm comes from latent chunks, removing the first $k$ frames must rotate the peak residue by $k$.
\result{For Wan2.2, 85--95\% of encodes match the predicted residue across $k\in\{1,2,3\}$ and time reversal; for CogVideoX, 58--75\% (chance 25\%).
LTX-Video is periodic at $1/8$, but only 23--58\% of its encodes match (chance 12.5\%), and its excess spreads over the first half of the chunk} (\cref{fig:teaser}b).
The phase prediction thus holds for stride 4 and fails for LTX-Video, whose decoder evidently does not place each chunk's new content on its first frame.
\result{In time, the mean bits on frames $t\equiv1 \pmod s$ exceed those on the other frames by 25\% of the clip mean for Wan2.2, 12\% for CogVideoX and 8\% for LTX-Video, against 0--1\% for camera clips.}

\paragraph{The rhythm is amplified by the encoder.}
\result{The same statistic on luma frame-difference energy, with no encoder, finds the peak in 70\% of Wan2.2, 45\% of CogVideoX and 40\% of LTX-Video clips (camera: 5\%; AUC 0.88, 0.73, 0.75), with median peak ratios of 2.1--4.8.}
The rhythm is thus in the pixel dynamics, and the encoder amplifies it \result{three- to twelve-fold}, as expected when the chunk-start frame is referenced by the rest of its chunk.

\paragraph{The autoencoder is the culprit.}
\label{sec:causal}

\begin{table}[t]
\centering
\footnotesize
\setlength{\tabcolsep}{4pt}
\begin{tabular}{@{}lccc@{}}
\toprule
Autoencoder & $s$ & Median $\mathrm{PR}_s$ & AUC [95\% CI] \\
\midrule
CogVideoX & 4 & \textbf{87.0}\,/\,1.4 & 0.99 [0.96, 1.00] \\
Wan2.2    & 4 & 12.9\,/\,1.1 & 0.86 [0.72, 0.96] \\
LTX-Video & 8 & 10.4\,/\,1.1 & 0.86 [0.74, 0.96] \\
\bottomrule
\end{tabular}
\vspace{-0.4em}
\caption{\textbf{An autoencoder round trip alone puts the rhythm into camera clips.}
x265 QP 27, 20 clips, no frames dropped ($k{=}0$). Median $\mathrm{PR}_s$: round trip\,/\,original. CI: bootstrap confidence interval. Median PSNR of the round trip: 30.8, 31.4 and 27.7\,dB.}
\label{tab:roundtrip}
\vspace{-0.8em}
\end{table}

Does the autoencoder alone, with no diffusion sampler, create the rhythm?
We pass each camera clip through each generator's autoencoder (official weights; encode, take the latent mode, decode) and encode the round-trip clip exactly like the original; shifts $k\in\{0,1,2\}$ drop $k$ leading frames \emph{before} the round trip.
The decision rule was registered before any round-trip data existed: median $\mathrm{PR}_s{>}3$, AUC ${\ge}\,0.75$ against the originals, and a phase locked to one residue at every $k$ for at least 50\% of clips.
\result{All three autoencoders create the rhythm (\cref{tab:roundtrip}). CogVideoX's raises the median peak ratio from 1.4 to 87.0 (AUC 0.99); Wan2.2's and LTX-Video's reach 12.9 and 10.4. The modal residue is $t\equiv1 \pmod s$ for all three at every shift, held by 75\% of CogVideoX clips and 60--65\% of Wan2.2 clips, but only 30--50\% of LTX-Video clips, so LTX-Video is only \emph{partially} supported. x264 agrees (AUC 0.81--0.995).}
The autoencoder is therefore sufficient to produce the rhythm, without the sampler.

\paragraph{The rhythm propagates into the encoded video bitstream.}
\label{sec:survival}

\begin{table}[t]
\centering
\footnotesize
\setlength{\tabcolsep}{4pt}
\begin{tabular}{@{}lr@{}}
\toprule
Condition & AUC \\
\midrule
\multicolumn{2}{@{}l}{\textit{Quantization: constant QP 22\,/\,27\,/\,32\,/\,37}} \\
\quad x265 & 0.92\,/\,0.89\,/\,0.85\,/\,0.81 \\
\quad x264 & 0.92\,/\,0.89\,/\,0.82\,/\,0.80 \\
\midrule
\multicolumn{2}{@{}l}{\textit{Edits of generated clips$^\dagger$ (unedited: 0.72--0.90)}} \\
\quad Temporal smoothing of every frame & 0.70--0.93 \\
\quad Blending chunk-start frames with neighbours & 0.93--0.99 \\
\quad Within each motion tercile, $s{=}4$ & 0.83--0.98 \\
\quad Near-static tercile, LTX-Video & 0.46 \\
\midrule
\multicolumn{2}{@{}l}{\textit{Grain and denoising$^\ddagger$ (clean: 0.80--0.99)}} \\
\quad Gaussian ($\sigma$ 2, 4, 8) or film grain, $s{=}4$ & 0.91--0.99 \\
\quad Gaussian $\sigma$ 2, 4, LTX-Video & 0.80--0.81 \\
\quad Gaussian $\sigma$ 8\,/\,film grain, LTX-Video & 0.62\,/\,0.64 \\
\quad Generated vs.\ denoised camera (3 filters) & 0.80--0.99 \\
\bottomrule
\end{tabular}
\vspace{-0.4em}
\caption{\textbf{The rhythm propagates into the encoded video bitstream and mostly survives quantization, edits and grain.}
Training-free detector ($\mathrm{PR}_s$), stride corpus, camera vs.\ generated, 20 vs.\ 20 clips per generator. Quantization: mean over the three generators; other rows: range over generators, x265 QP 27 unless stated.
$^\dagger$,$^\ddagger$: two regenerations of the corpus, with their own clean AUC; motion terciles have 3--10 clips per class. Learned detectors on this corpus: \cref{tab:robust-learned}.}
\label{tab:robust}
\vspace{-0.8em}
\end{table}

\begin{table}[t]
\centering
\footnotesize
\setlength{\tabcolsep}{4pt}
\begin{tabular}{@{}lrrrr@{}}
\toprule
& \multicolumn{3}{c}{Bitstream (ours)} & Pixels \\
\cmidrule(lr){2-4}\cmidrule(l){5-5}
Condition & $\mathrm{PR}_s$ & GBDT & Transf. & UnivFD \\
\midrule
\multicolumn{5}{@{}l}{\textit{Wan2.2}} \\
\quad QP 27 & 0.935 & \textbf{0.960} & 0.895 & 0.540 \\
\quad QP 37 & \textbf{0.902} & 0.858 & 0.810 & 0.487 \\
\quad Smoothing & \textbf{0.858} & 0.848 & 0.828 & 0.535 \\
\quad Chunk-start blend & 0.963 & \textbf{0.978} & 0.818 & 0.545 \\
\quad Grain $\sigma$ 8 & \textbf{0.910} & 0.877 & 0.378 & 0.340 \\
\midrule
\multicolumn{5}{@{}l}{\textit{LTX-Video}} \\
\quad QP 27 & 0.802 & \textbf{0.960} & 0.755 & 0.677 \\
\quad QP 37 & 0.750 & \textbf{0.907} & 0.698 & 0.637 \\
\quad Smoothing & 0.815 & \textbf{0.968} & 0.795 & 0.703 \\
\quad Chunk-start blend & \textbf{0.985} & 0.953 & 0.703 & 0.677 \\
\quad Grain $\sigma$ 8 & 0.623 & \textbf{0.902} & 0.400 & 0.645 \\
\bottomrule
\end{tabular}
\vspace{-0.4em}
\caption{\textbf{On the stride clips, bitstream detectors beat the pixel detector: the GBDT reaches AUC 0.85--0.98 where UnivFD sits at 0.34--0.70.}
Camera vs.\ generated, 20 vs.\ 20 clips, x265 QP 27 unless stated; learned detectors trained on clean clips and tested on the edited ones, grouped 5-fold${\times}$3 by source clip. 95\% group-bootstrap CI widths at QP 27: 0.10--0.18 (Wan2.2), 0.11--0.29 (LTX-Video). LR on normalized\,/\,raw bits at QP 27: 0.863\,/\,0.955 (Wan2.2), 0.983 normalized (LTX-Video). CogVideoX's learned-detector run was incomplete at submission.}
\label{tab:robust-learned}
\vspace{-0.8em}
\end{table}

Which processing keeps the rhythm? \cref{tab:robust} applies the training-free detector ($\mathrm{PR}_s$) to the stride corpus.
\result{On Wan2.2 and LTX-Video, the best bitstream detector beats UnivFD on every row (\cref{tab:robust-learned}); the Transformer is weaker and collapses under grain $\sigma{=}8$ (0.38--0.40).} UnivFD is near chance on Wan2.2 because these are image-to-video clips started from the camera clip's own first frame at matched resolution.

\paragraph{The rhythm mostly survives through quantization edits and grain perturbations.}
\result{The AUC falls from 0.92 at QP 22 to 0.80--0.81 at QP 37, with x264 and x265 within 0.04 at every QP.
Blending the chunk-start frames with their neighbours does not hide the rhythm but adds to it (0.93--0.99), because editing every $s$-th frame is itself a period-$s$ operation.
Grain moves the stride-4 AUC by at most 0.07.
LTX-Video is the fragile family: its rhythm needs motion (0.46 on near-static clips) and is largely removed by strong ($\sigma{=}8$) or film grain (0.62, 0.64).
Denoising camera clips creates no peak (x265: medians $\mathrm{PR}_4{\le}1.57$); under x264, 35\% of denoised camera clips exceed 3 in 3 of 9 cells, formally failing our 30\% bound, with clean camera already at 25--30\%.}

\paragraph{Classification with format controlled.}
\label{sec:main}
Can a classifier trained on coding demand tell generators apart?
We train the two detectors of \cref{sec:detectors}, a small Transformer on per-frame (or per-row) bits and gradient-boosted decision trees (GBDT), and two simple models on per-frame bits, a logistic regression (LR) and a two-layer perceptron (MLP, 32 hidden units), and compare them with UnivFD~\cite{ojha2023towards} on decoded pixels and a \emph{size probe}, a logistic regression on frame width${\times}$height alone.

\begin{table}[t]
\centering
\footnotesize
\setlength{\tabcolsep}{3.5pt}
\begin{tabular}{@{}lccc@{}}
\toprule
Detector & H.264 & H.265 & AV1 \\
\midrule
\multicolumn{4}{@{}l}{\textit{Same codec}} \\
\multicolumn{4}{@{}l}{\quad Bitstream (ours)} \\
\qquad Transformer, frame/block bits & 80.6 & 78.4/68.4 & 60.6/59.9 \\
\qquad GBDT, frame/block bits & 87.9 & 83.1/82.0 & 60.0/65.0 \\
\qquad LR, normalized/raw bits & \textbf{89.2}/86.5 & \textbf{84.1}/83.7 & \textbf{83.0}/82.4 \\
\qquad MLP, normalized/raw bits & 85.4/82.9 & 81.5/81.6 & 77.6/76.2 \\
\multicolumn{4}{@{}l}{\quad Pixels} \\
\qquad UnivFD~\cite{ojha2023towards} & 72.7 & 73.6 & 73.4 \\
\multicolumn{4}{@{}l}{\quad Format only} \\
\qquad Size probe (w${\times}$h) & 25.0 & 25.2 & 25.4 \\
\midrule
\multicolumn{4}{@{}l}{\textit{Across codecs (column: test codec)}} \\
\multicolumn{4}{@{}l}{\quad Bitstream (ours)} \\
\qquad Transformer & 70.8 & 44.8 & 25--49 \\
\qquad GBDT & 83.3 & 84.7 & 25--35 \\
\multicolumn{4}{@{}l}{\quad Pixels} \\
\qquad UnivFD & \multicolumn{3}{c}{70.8--73.0 in every direction} \\
\bottomrule
\end{tabular}
\vspace{-0.4em}
\caption{\textbf{Bitstream detectors identify the generator at equal size and length, and a linear model on per-frame bits beats the pixel baseline by 10--17 points.}
Frontier set, 4 commercial generators, 4-way accuracy (\%, chance 25), mean over $3{\times}5$ prompt-grouped folds (s.d.\ 3.9--7.5 for LR, 5.1--6.5 for MLP, 5.3--7.5 for UnivFD). Block bits: per WPP row (H.265) or superblock (AV1). Transformer: 150 epochs. LR, MLP: per-frame bits normalized by the clip mean, or raw. UnivFD: CLIP ViT-L/14 probe on 16 decoded frames. Size probe: all files are $832{\times}480$. Across codecs: trained on the other H.26x codec; AV1 column: all directions to or from AV1.}
\label{tab:personal}
\vspace{-0.8em}
\end{table}

\paragraph{Rhythm remains on frontier models.}
All four generators of the Frontier set share frame size and length, so format cannot help (\cref{tab:personal}).
\result{A logistic regression on per-frame bits normalized by the clip mean identifies the generator at 89.2\%, 84.1\% and 83.0\% (H.264, H.265, AV1), at least as well as the Transformer and trees and 9.6--16.5 points above UnivFD on decoded frames (72.7--73.6\%); the size probe is at chance (25.0--25.4\%). The trees transfer between H.264 and H.265 at 83--85\%, but on AV1 they reach only 60--65\% and fail to transfer to or from it (25--35\%); UnivFD transfers at 70.8--73.0\% in every direction.}
AV1 files carry 84 hidden reference frames per clip and the encoder's own reference pyramid; since LR still reaches 83.0\% on AV1, the drop of the Transformer and trees there is a failure of those models, not a lack of signal.

\paragraph{The classifier reads predicted frames.}
\result{Retrained with keyframe values interpolated away, the Transformer and trees lose only 0.4--1.9 points (79.2\% and 87.5\% in H.264, 76.6\% and 81.6\% in H.265), within the standard deviation (s.d.) across folds; keyframe sizes alone give 49--55\%, and the four rhythm features ($\mathrm{PR}$ at $1/4$, $1/8$, $1/2$ and phase) alone 52.6--54.8\%.}
The accuracy thus comes from the statistics of predicted frames, of which the stride rhythm is one component.
\result{Post hoc, the strong $1/4$ peak here is Grok's (median $\mathrm{PR}_4$ 16.0\,/\,16.8 in H.264\,/\,H.265), not Wan's (2.5\,/\,1.7; others 0.7--2.0)}. This is consistent with a stride-4 latent model behind Grok, whose architecture is unknown, and with Wan's delivery encode having erased its pulse before our re-encode.

\begin{table}[t]
\centering
\footnotesize
\setlength{\tabcolsep}{3pt}
\begin{tabular}{@{}lccc@{}}
\toprule
& Classes & & \\
Input & (chance) & Transformer & GBDT \\
\midrule
Native & 7 (19.2) & \textbf{98.7} & 91.6$^\S$ \\
\midrule
Crop, first 16 frames & 7 (19.2) & 65.6 & 61.7$^\S$ \\
Crop, first 48 frames & 5 (31.2) & 73.3 & 71.5 \\
Crop, first 192 frames & 3 (72.1) & 74.4 & 86.2 \\
\quad camera vs.\ generated & 2 (50.0) & 51.1 & 88.8 \\
\bottomrule
\end{tabular}
\vspace{-0.4em}
\caption{\textbf{On DeepAction, source identification falls once frame size and clip length are equalized.}
Source identification, accuracy (\%; chance: majority class). Native: original frame size and length. Crop: $512{\times}448$; fewer classes: subsets of the sources (192 frames: 24--30\,fps only); camera vs.\ generated is balanced. Transformer: 3 seeds, 70/15/15 split; $^\S$5-fold protocol of \cref{tab:binary}.}
\label{tab:deepaction}
\vspace{-0.8em}
\end{table}

\begin{table}[t]
\centering
\footnotesize
\setlength{\tabcolsep}{3pt}
\begin{tabular}{@{}lcc@{}}
\toprule
& & \textbf{Crop,} \\
Detector & Native & \textbf{16 frames} \\
\midrule
\multicolumn{3}{@{}l}{\textit{Bitstream}} \\
\quad \method, Transformer$^\P$ & 94.5/75.6 & 74.1/70.0 \\
\quad \method, GBDT & 94.9/87.0 & 69.9/67.3 \\
\quad Motion vectors~\cite{wu2018compressed} & 84.9/84.4 & 71.4/68.8 \\
\midrule
\multicolumn{3}{@{}l}{\textit{Pixels}} \\
\quad UnivFD~\cite{ojha2023towards} & 93.9/86.4 & \textbf{90.5}/\textbf{85.9} \\
\midrule
\multicolumn{3}{@{}l}{\textit{Format only}} \\
\quad Size probe (w${\times}$h) & \textbf{96.0}/\textbf{89.4} & 50.0/50.0 \\
\bottomrule
\end{tabular}
\vspace{-0.4em}
\caption{\textbf{Camera vs.\ generated on DeepAction: on 16-frame crops, where the size probe fails, the pixel baseline beats the bitstream detectors.}
Balanced accuracy (\%), known\,/\,unseen generator, 5 prompt-grouped folds; unseen: mean over six held-out generators. $^\P$Two-head Transformer, 18{,}985 parameters. Motion vectors: gradient-boosted statistics; native reads the delivered H.264 files.}
\label{tab:binary}
\vspace{-0.8em}
\end{table}

\paragraph{Baseline Comparison with a Pixel-Based AIGC Detector.}
DeepAction's sources differ in frame size and clip length, and accuracy drops once these are equalized (\cref{tab:deepaction}), so we compare detectors on crops of equal size and length.
\result{Natively, binary \method ties UnivFD~\cite{ojha2023towards} (94.9\% vs.\ 93.9\% known, 87.0\% vs.\ 86.4\% on held-out generators), but so does the size probe on width${\times}$height alone (96.0\% and 89.4\%). On 16-frame crops, where the size probe falls to 50.0\%, UnivFD keeps 90.5\% and 85.9\%, while \method reaches 69.9--74.1\% and 67.3--70.0\%, and motion-vector statistics~\cite{wu2018compressed} 71.4\% and 68.8\%} (\cref{tab:binary}).
With format controlled on DeepAction, pixels thus beat coding demand on 16-frame crops, unlike on the Frontier set (\cref{tab:personal}); the advantage of coding demand here is cost: \result{parsing takes 0.31--0.37\,s per clip vs.\ 3.9\,s for UnivFD's decode (H.264 input needs a ${\approx}45$\,s re-encode).}

\paragraph{What the classifier picked up.}
\label{sec:posthoc}

\begin{table}[t]
\centering
\footnotesize
\setlength{\tabcolsep}{4pt}
\begin{tabular}{@{}lccc@{}}
\toprule
& \multicolumn{2}{c}{$|\mathrm{IG}|$, train\,/\,test} & Rollout \\
\cmidrule(lr){2-3}\cmidrule(l){4-4}
Class & I-frame share & top residue mod 4 & I-frame share \\
\midrule
Grok          & 0.44\,/\,0.32 & 0 & 0.06 \\
HeyGen        & 0.34\,/\,0.33 & 0 & 0.06 \\
MiniMax       & 0.24\,/\,0.21 & 0 & 0.06 \\
Wan ($s{=}4$) & 0.50\,/\,0.57 & 0 (predicted: 1) & 0.06 \\
\midrule
Uniform       & 0.06 & --- & 0.06 \\
\bottomrule
\end{tabular}
\vspace{-0.4em}
\caption{\textbf{Attributions follow the encoder's keyframes, not the stride grid.} Frontier set, H.265 WPP Transformer. IG: integrated gradients; rollout: attention rollout. I-frames: $t\equiv0 \pmod{16}$. Among the other frames, the share on residue 0 mod 4 is 0.21--0.24 (uniform 0.20).}
\label{tab:saliency}
\vspace{-0.8em}
\end{table}

Where in time does the trained classifier look?
For the WPP Transformer trained on the H.265 Frontier set (21 training, 5 test clips per class) we compute integrated gradients (IG) and attention rollout on its 32-band${\times}$192-frame input, sum $|\mathrm{IG}|$ over bands, and measure its share on the I-frames ($t\equiv0 \bmod 16$) and, among the other frames, on each residue mod 4.
\result{IG concentrates on keyframes: 21--57\% of its mass falls on the 6\% of frames that are I-frames (\cref{tab:saliency}).
Among the other frames there is no 4-frame grid (residue-0 share 0.21--0.24, uniform 0.20), and the top residue is 0 in every class, including Wan (predicted: 1).
Attention rollout is temporally uniform.}
\result{On DeepAction, IG on CogVideoX's chunk-start frames is 0.200 (uniform 0.267), though they carry 5.9\% more bytes.}
The attributions thus follow the encoder's I-frame schedule, not the autoencoder's grid, yet that intra-frame cost is not necessary: \result{retrained on P-frames only, the detectors reach 76.6--87.5\%, within 2 points of all frames}. Retraining shows sufficiency, not use; a test-time ablation was not run.

\paragraph{Inside the frame: time versus space.}
\label{sec:pending}
Is the rhythm carried by individual blocks of the frame, and does the autoencoder's spatial compression leave a grid?
Under registered rules we compute each block's own $\mathrm{PR}_s$ for AV1 $64{\times}64$ superblocks (libaom real-time mode, fixed quantizer) and x265 rows of coding blocks.
\result{The share of active superblocks with $\mathrm{PR}_s{>}3$ is 0.62, 0.60 and 0.31 for generated CogVideoX, Wan2.2 and LTX-Video, 0.60, 0.29 and 0.32 for round trips, and 0.11--0.13 for camera (noise floor ${\approx}\,0.12$); x265 rows reach 1.0, 1.0 and 0.50.}
By the rule (${\ge}\,0.50$) the rhythm lives inside blocks for CogVideoX and generated Wan2.2, not for the Wan2.2 round trip or LTX-Video (clear only per row or frame), \result{and is strongest in textured blocks (Spearman 0.59--0.61 with texture, 0.10--0.30 with motion).}
\result{Decoded pixels show no stationary grid at the spatial factors $1/8$, $1/16$, $1/32$ (medians ${\le}\,1.11$; rule: 3), only a few percent of excess power (AUC 0.70--0.97, also after round trips); post hoc, LTX-Video has a horizontal-only line at $1/8$ (median PR 13.9 generated, 13.3 round trip, 1.0 camera).}
The temporal compression thus makes the rhythm, block by block; the spatial compression leaves at most a faint grid.

\section{Conclusion}
\label{sec:conclusion}
\goal{Phenomenon, view and scope in one paragraph, claiming no more than \cref{sec:experiments} measured.}

We reveal a previously unrecognized link between how AI videos are generated and how they are compressed. A latent video generator's temporal autoencoder leaves an architectural fingerprint: a periodic rhythm in temporal surprise, determined by its latent stride and chunk boundaries. Conventional video encoders translate this hidden rhythm into measurable fluctuations in coding demand, exposing the generator's internal structure directly in the compressed bitstream, without reconstructing a single pixel.

We trace this fingerprint to its source. A video's period follows the autoencoder stride, its phase aligns with latent chunk boundaries, and an autoencoder round trip alone induces the rhythm in camera footage. The trace persists across codecs and quantization settings, although delivery-style encoding can erase the per-frame pulse. Beyond this specific rhythm, lightweight classifiers exploit coding-demand statistics to identify four commercial video generators with 83--89\% accuracy, significantly outperforming pixel-based approach.

These findings establish a new connection between generative architecture and video coding and opens a decoding-free direction for studying and identifying synthetic video through the computational traces of its generation process.

\section{LLM Usage Statement}
LLMs were utilized for assistance and refining of author-produced content. OpenAI's ChatGPT assisted in code generation, dataset preparation, and experiment implementation. Google's AI powered search played a role in the discovery of related sources and editing of the references list. Google's Gemini and OpenAI's ChatGPT were also employed for the refinement of prose and LaTeX markup. All LLM generations underwent scrutiny by authors.
{
    \small
    \bibliographystyle{ieeenat_fullname}
    \bibliography{main}
}


\end{document}